\documentclass[sigconf]{acmart}

\usepackage{graphicx}
\usepackage{xcolor}
\usepackage[ruled,vlined,linesnumbered]{algorithm2e}
\usepackage{multirow}
\usepackage{dblfloatfix} 
\usepackage{pifont}

\AtBeginDocument{%
  \providecommand\BibTeX{{%
    \normalfont B\kern-0.5em{\scshape i\kern-0.25em b}\kern-0.8em\TeX}}}

\acmConference[Milan '22]{ACM/IEEE International Conference on Cyber-Physical Systems}{May 2022}{Milan, Italy}%

\begin{document}

\title{\textsc{HydraFusion}: Context-Aware Selective Sensor Fusion for Robust and Efficient Autonomous Vehicle Perception}



\author{Arnav Vaibhav Malawade}
\authornote{Both authors contributed equally to this research.}
\author{Trier Mortlock}
\authornotemark[1]
\author{Mohammad Abdullah Al Faruque}
\affiliation{%
  \institution{University of California, Irvine}
  \city{Irvine}
  \state{California}
  \country{USA}
  }

\renewcommand{\shortauthors}{Malawade and Mortlock, et al.}

\begin{abstract}
Although autonomous vehicles (AVs) are expected to revolutionize transportation, robust perception across a wide range of driving contexts remains a significant challenge. Techniques to fuse sensor data from camera, radar, and lidar sensors have been proposed to improve AV perception. However, existing methods are insufficiently robust in difficult driving contexts (\textit{e.g.}, bad weather, low light, sensor obstruction) due to rigidity in their fusion implementations. These methods fall into two broad categories: (i) early fusion, which fails when sensor data is noisy or obscured, and (ii) late fusion, which cannot leverage features from multiple sensors and thus produces worse estimates. To address these limitations, we propose \textbf{\textsc{HydraFusion}}: a selective sensor fusion framework that learns to identify the current driving context and fuses the best combination of sensors to maximize robustness without compromising efficiency. \textbf{\textsc{HydraFusion}} is the first approach to propose dynamically adjusting between early fusion, late fusion, and combinations in-between, thus varying both \textit{how} and \textit{when} fusion is applied. We show that, on average, \textbf{\textsc{HydraFusion}} outperforms early and late fusion approaches by \textbf{13.66\%} and \textbf{14.54\%}, respectively, without increasing computational complexity or energy consumption on the industry-standard Nvidia Drive PX2 AV hardware platform. We also propose and evaluate both static and deep-learning-based context identification strategies. Our open-source code and model implementation are available at \url{https://github.com/AICPS/hydrafusion}.

\end{abstract}





\maketitle

\section{Introduction}
Autonomous vehicles (AVs) are cyber-physical systems (CPSs) that operate in complex, dynamic environments with many different actors. An AV must be able to perceive the environment accurately and efficiently to ensure safety across driving settings. Most modern AVs are equipped with multiple sensors and use sensor fusion techniques to help handle the uncertainties present in challenging driving scenes. 
Even with these methods, autonomous driving is a highly complex task, and large deep-learning algorithms are necessary to enable accurate perception. 

Despite recent advances, industry-standard AV perception systems still tend to fail in difficult contexts \cite{NTSB2019uber, NTSB2018}. A na\"ive solution to the problem is to continue increasing the size and complexity of AV algorithms and incorporate more sensors to cover as many driving contexts as possible. However, AVs are energy-constrained CPSs, so the use of larger algorithms comes at the cost of reduced driving range, increased expense, and increased power and thermal demands on the vehicle \cite{lin2018architectural}. Moreover, as shown in Section \ref{subsec:motivation}, in some contexts fusing \textit{more} sensors can actually result in a \textit{less} precise result.
Thus, robust and accurate AV perception requires algorithms that can \textit{adapt} to dynamically changing driving contexts as they appear without increasing the computation requirements.

Typical AV perception systems implement deep convolutional neural networks (CNNs) \cite{ren2015faster}, in which sensor measurements are fed through a series of convolutional layers to produce spatial features. These features are then used to detect objects in different regions of the visual scene. 
Sensor performance can vary depending on factors such as weather, lighting, and physical obstructions \cite{shahian2019real,rosique2019systematic,sheeny2020radiate}. Sensor fusion algorithms attempt to combine the benefits from each sensor to produce a more accurate result. However, in dynamic environments, the \textit{context} of the scene is often overlooked or excluded from the fusion method entirely. 
Most modern multi-sensor approaches typically perform sensor fusion at only one point in the model, whether it be early fusion across the raw sensor measurements or late fusion after detections have been made \cite{nobis2019deep,shahian2019real,xu2018pointfusion}. Furthermore, most works use static algorithms for fusion that do not depend on the context of the AV's operating environment. 
Context-aware sensing approaches have proven beneficial for a wide range of CPS applications \cite{ivanov2018context,fong2020contextually}.
Humans intuitively leverage contextual information about the driving scene (\textit{e.g.}, weather, lighting, road type, high-level visual features) to adjust their decisions and focus while driving.
Similarly, contextual information can inform AV perception and enable more robust fusion in complex driving contexts.

The scope of this paper addresses the following core research problems: (i) implementing a fusion approach that is robust across diverse contexts, noise sources, and sensor error types; (ii) using the context of a scene to improve sensor fusion performance; and (iii) implementing an efficient multi-sensor fusion approach for energy-constrained AV edge devices. 

In this paper, we propose \textsc{HydraFusion} --- a context-aware sensor fusion approach that actively identifies the driving context and uses it to selectively fuse sensor data from different modalities at varying depths in the model. 
By using a selective sensor fusion approach, \textsc{HydraFusion} can improve the robustness of AV perception without increasing the computational demands on the energy-constrained AV edge platform. 
Our work is the first to study a context-aware selective sensor fusion approach that can dynamically adjust both \textit{how} and \textit{when} fusion is applied. We specifically study the problem of object detection in the AV perception system; however, we posit that our proposed approach can be applied to a variety of cyber-physical sensor fusion applications, including tracking, localization, and mapping \cite{chen2019selectfusion,gokhale2021feel,saeedi2014context,marti2012context}. 
The key contributions of this work are as follows:
\begin{enumerate}
    \item We propose a novel multi-branch sensor fusion architecture that enables early fusion, late fusion, as well as intermediate combinations. 
    \item We propose intelligent, context-aware gating strategies that maximize robustness by dynamically selecting the fusion methodology depending on the current context.  
    \item We demonstrate that our approach outperforms existing methods on a challenging real-world dataset containing a wide range of driving contexts, including bad weather, poor lighting, and various location types.
    \item We implement our approach on an industry-standard AV hardware platform, the Nvidia Drive PX2, to demonstrate that our approach can be practically deployed in a real AV with comparable energy consumption, latency, and memory usage to state-of-the-art methods.
    \item We open-source our algorithmic implementation and architecture\footnote{\url{https://github.com/AICPS/hydrafusion}} to benefit the research community and enable further study of selective sensor fusion approaches for CPS problems.
\end{enumerate}

In the remainder of this section, we provide a motivating example for our work. In Section \ref{sec:relatedwork}, we discuss related work. Section \ref{sec:formulation} presents our problem formulation. In Section \ref{methodology}, we discuss our proposed approach, \textsc{HydraFusion}, and in Section \ref{exp}, we provide numerical results on the performance of our approach. Concluding remarks are given in Section \ref{sec:conclusion}.



\subsection{Motivational Example}
\label{subsec:motivation}







\begin{figure*}[!ht]
    \centering
    \includegraphics[clip, trim= 45 180 85 40, width=1.0\linewidth]{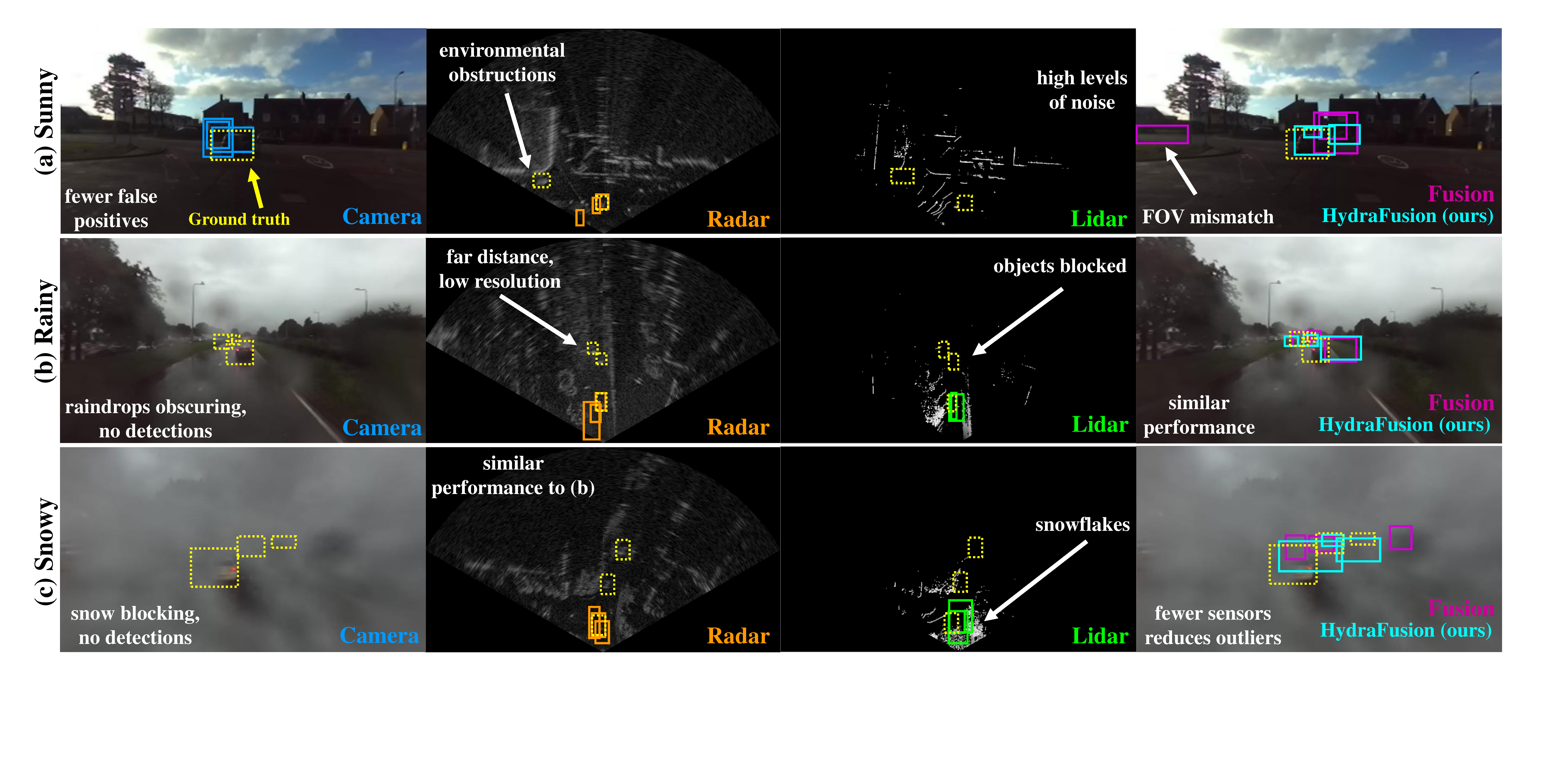}
    \caption{Qualitative analysis of object detection with different sensors and their fusion across three contexts. Ground truth detections are shown in yellow, while sensor-specific and fusion detections are shown in their respective colors. \textsc{HydraFusion} achieves the most accurate predictions across contexts.}
    \vspace*{-3mm}
    \label{fig:mot}
\end{figure*}

Here, we outline theory regarding the foundation of our approach. Then, to illuminate the diverse conditions an AV may encounter, we qualitatively analyze scenes from a public dataset. This section supplements the quantitative results we present in Section \ref{exp}.


\textbf{\textit{Theoretical Analysis.}} Our approach involves the fusion of multiple measurements to produce the most accurate estimate possible. From a generic standpoint, examining one measurement, we aim to model the relationship between the parameter we want to estimate and that measurement. This can be achieved through a commonplace parameter estimation formulation \cite{bar2004estimation} as follows:
\begin{equation}
    x = Hy + e  
\end{equation}
where $y$ is the true parameter we are attempting to estimate of dimension $n_y$, $x$ is the measurements to be fused of dimension $n_x$, $H$ is an $n_x \times n_y$ measurement matrix that provides a mapping between the state of the system to the measurements, and $e$ is the estimation error. The goal is to minimize the squared errors with respect to $y$. Thus, the objective can be defined as:
\begin{equation}
   \min_y||x-Hy||^2 
\end{equation}
After the partial derivative with respect to $y$ is taken and set to $0$, the least-squares solution is achieved\footnote{We assume that the required inverses exist throughout this section.}:
\begin{equation}
    \hat{y} = (H^TH)^{(-1)}H^Tx
\end{equation}
However, if we model the error, $e$, such that it is zero mean, $\mathbb{E}(e) = 0$ with $\mathbb{E}(ee') = R$ where $R$ is the error covariance, then the solution yields a weighted least-squares by the noise covariance matrix and can be derived as:
\begin{equation}
    \hat{y} = (H^TR^{-1}H)^{-1}H^TR^{-1}x
\end{equation}
Now, we consider the case of fusing $n$ sensor measurements. The minimum variance unbiased estimator can be derived in a batch manner as:
\begin{equation}
    \hat{y}^{(n)} = [\sum_{i=1}^{n} H_i^TR_i^{-1}H_i ]^{-1} \cdot \sum_{i=1}^{n} H_i^TR_i^{-1}x_i
    \label{simp}
\end{equation}


Typically, the more information a fusion filter has, the better the performance will be. However, this argument breaks apart in cases where there are discrepancies in the measurement models, among other sources of estimation errors. These conditions commonly occur in autonomous driving \cite{feng2020review} and can negatively impact the quality of sensor measurements. Given incorrect $R$ values in Eq. \ref{simp}, this can lead to known problems of model divergence and/or filter inconsistencies \cite{bar2004estimation}. For example, suppose a camera sensor on an AV has raindrops obscuring the lens. In that case, it may generate overconfident estimates where the $R$ value does not reflect the true amount of noise in the measurement. In this paper, we show that in certain scenarios involving sensor fusion in cyber-physical systems, it is \textit{not} optimal to fuse all the sensor measurements available and can actually \textit{reduce} estimation accuracy.

\textbf{\textit{Qualitative Analysis.}}
To illustrate these points, Figure \ref{fig:mot} visualizes the object detection results for a variety of contexts from a public driving dataset \cite{sheeny2020radiate}. 
Measurements from three sensing modalities (camera, radar, lidar) are shown from left to right across three different driving contexts: (a) \textit{sunny}, (b) \textit{rainy}, (c) \textit{snowy}. The ground truth objects in the scenes are shown in dotted yellow boxes. A deep-learning-based object detection pipeline featuring Faster R-CNN \cite{ren2015faster} with a ResNet-18 \cite{he2016deep} backbone was used to generate the detections for each sensor input. The fusion method, shown in purple in the last column for each scene, represents the standard approach to fuse detections from all sensors. In the same column, we also show our approach, \textsc{HydraFusion}, that selectively fuses sensors based on the context derived from a scene. For clarity, only the highest scoring predictions for each configuration are shown in the figure.

Some clear trends emerge in the results: (i) cameras predict fewer false positives but struggle in severe weather, as shown with the rainy and snowy images where the camera lens is obscured; (ii) radars can struggle in scenes with many objects blocking or deflecting measurements as shown in the urban setting, but remains robust in adverse weather conditions of rain and snow; and (iii) lidar can experience high levels of noise in a densely populated scene, can miss objects that are behind other objects, and can degrade in performance due to weather like the snowflakes shown in the figure. A summary of each modality's qualitative performance in different contexts of the dataset is shown in Table \ref{tab:mot_tab}.  

\begin{table}[htb]
\vspace{0.5mm}
    \centering
    \begin{tabular}{c c c c c}
    \hline
        Scene & Camera & Radar & Lidar & Fusion\\\hline
        {Urban} & {\checkmark} & {\ding{53}} & {\ding{53}} & {\checkmark}\\\hline
        {Rainy} & {\ding{53}} & {\checkmark} & {\checkmark} & {\checkmark}\\\hline
        {Foggy} & {\ding{53}} & {\checkmark} & {\checkmark} & {\checkmark}\\\hline
        {Snowy} & {\ding{53}} & {\checkmark} & {\ding{53}} & {\checkmark}\\\hline
        {Night} & {\ding{53}} & {\checkmark} & {\checkmark} & {\checkmark}\\\hline
    \end{tabular}
    \caption{Qualitative object detection performance of each sensing modality in different driving contexts.}
    \label{tab:mot_tab}
    \vspace{-6mm}
\end{table}
The last column of Figure \ref{fig:mot} allows us to examine fusion across the three different scenes. For most objects, the fusion method performs better than a single sensing modality. However, there are specific drawbacks to this approach. In (a), a field-of-view (FOV) mismatch arises between fusing detections across different modalities. Furthermore, the original camera detections for the center object were skewed far to the right by the other sensors' predictions. In (c), it is clear that the fusion method predicts more outliers that deviate from the ground truth.  Overall, a more optimal estimate across all the images is achieved by fusing only a subset of sensors, as done in our approach, \textsc{HydraFusion}. This result motivates the need for a selective sensor fusion approach that can dynamically adjust to different contexts. The experimental results shown in Section \ref{exp} of this paper further validate the theoretical and qualitative analysis provided here. 







\section{Related Work}
\label{sec:relatedwork}
This section discusses related works on sensor fusion, object detection, and multi-branch deep learning. We elaborate on their scope and limitations and compare them with our proposed approach.

\subsection{Sensor Fusion} 
In traditional sensor fusion approaches that have known dynamics, noise, and measurement models, more sensors can help achieve better results \cite{bar2004estimation}. 
Fusion across multiple homogeneous sensors can help reduce uncertainties by increasing \textit{confidence} or providing measurements over a wider observation area to increase \textit{coverage}. 
Fusing heterogeneous sensors can also reduce sensing uncertainties by providing information across a different feature set for the same task.
However, the fusion of all sensors does not always guarantee better estimates, especially with highly nonlinear and dynamic systems such as AV perception systems. Hence, there are potential benefits to selectively fusing information obtained from sensors, as shown in some recent works. In \cite{chen2019selective}, a selective sensor fusion scheme is developed for a visual-inertial odometry system to provide robustness against data corruption. The authors implement feature selection using data-driven models that consider measurement reliability and vehicle-environment dynamics. This work is extended to a generic framework for selective sensor fusion in deep pose estimation in \cite{chen2019selectfusion}. However, these works only implement late-fusion over the outputs of sensor-specific deep learning models, limiting their performance and efficiency. 
Authors in \cite{lee2020accuracy} propose a strategy to alter the power levels and operating state of an AV lidar sensor depending on the vehicle’s speed and environment. Similarly, \cite{gokhale2021feel} proposes adjusting the sensing frequency for indoor robot localization according to environmental dynamics. These approaches primarily focus on improving sensor efficiency.
In contrast to these related works, our approach is the first to propose selective fusion for AVs with a dynamic gating component. By selecting between multiple modalities and fusion locations, our approach maximizes robustness by selecting both \textit{how} and \textit{when} fusion takes place in the model.

In a similar vein, several works have studied the use of contextual information from the environment within an information fusion framework. 
Authors in \cite{snidaro2015context} survey context-based information fusion and discuss how different types of contextual information interact with state variables and traditional fusion approaches. 
Both \cite{saeedi2014context} and \cite{marti2012context} show that context-aided sensor fusion frameworks for navigation improve robustness over standard methods. 
Distinct from these works, our approach utilizes deep learning models to learn contextual representations of scenes instead of static fusion rules to provide more robust results. 
Authors in \cite{gong2019context} extract contextual information using specialized feature mining within a CNN for object detection in very-high-resolution imagery. However, their approach is focused on obtaining contextual information from regions of interest in images, whereas our approach extracts the context of a \textit{scene} using multiple heterogeneous sensory inputs. 

\subsection{Fusion in Object Detection Methods} 
Traditional object detection methods use CNNs to extract spatial features from inputs to identify objects in the scene \cite{ren2015faster}.
Object detection in AVs is more challenging as the physical aspects of the environment affect performance. Both \cite{feng2020review} and \cite{arnold2019survey} survey object detection in AVs;
\cite{feng2020review} focuses on probabilistic methods, while \cite{arnold2019survey} studies 3D detection methods. Both papers identify gaps in modeling sensor uncertainty. As detailed in the previous subsection, sensor fusion methods can help offset some measurement inaccuracies. 


Fusion methods in object detection largely fall into two main categories: feature-level (or \textit{early}) fusion and decision-level (or \textit{late}) fusion. Early fusion approaches can extract many multi-modal features from the input but can be sensitive to noise and outliers from the sensors, reducing their robustness \cite{nobis2019deep,shahian2019real}. Late fusion methods are more robust to sensor noise but cannot combine intermediate features across sensors, limiting their performance \cite{xu2018pointfusion}.
%
\textsc{HydraFusion} remains unique in its approach of combining early and late fusion approaches. To the best of our knowledge, this is the first work to propose a multi-layered fusion approach for object detection in AV perception systems. 
\subsection{Multi-Branch Architectures in Deep Learning}
\textsc{HydraFusion} maintains computational efficiency when evaluating multiple object detection pipelines simultaneously by utilizing a gating strategy, which limits the number of detection pipelines, or \textit{branches}, that are run. 
Several types of multi-branch deep learning approaches have been proposed for image processing tasks. 
In \cite{ahmed2016network}, a network of experts approach to image categorization is proposed. Each branch is a CNN that only discriminates between the subset of classes it is assigned to learn, as this approach lacks an intelligent gating module. Similarly, \cite{aljundi2017expert} uses specific expert branches but focuses on life-long learning and the generation of new tasks and experts.

\cite{mullapudi2018hydranets} explores efficient methods for single image classification, where the authors use branches developed to compute features on visually similar classes.
During training, the authors employ an adaptive form of dropout where entire branches are dropped when they are not chosen by the gating function. 
Similarly, TridentNet \cite{li2019scale} is a network that addresses the problem of scale variation in object detection. Its three-branch architecture shares parameters and structure between branches, resulting in faster training and inference and the enforcement of similar operations across feature maps, but this requires similarly structured branches. 
Our approach fundamentally differs from these works in that \textsc{HydraFusion} takes in multiple heterogeneous sensor modalities as inputs, incorporates context into an intelligent branch selection method, and targets dynamic sensor fusion for robust object detection via a multi-branch approach. 
Our approach is also unique because it enables the specialization of branches to individual sensors or subsets of sensors to improve robustness across varying driving contexts. 

\section{Problem Formulation}
\label{sec:formulation}
This section provides a formulation for the key sensor fusion problem targeted in this paper: object detection in AVs. We assume that the AV uses a variety of sensing modalities to take measurements of the driving scene. At discrete time steps, samples are generated, which consist of input measurements, $\mathbf{X}$, from the sensors. 
The objective is to accurately detect objects, $\mathbf{Y}$, within each scene using the sensor measurements:
\begin{equation}
    \mathbf{Y} = \phi(\mathbf{X}) ,  
\end{equation}
\begin{equation}
    \mathbf{Y} = \{\mathbf{Y}_{class}^i, \mathbf{Y}_{reg}^i\}_{i=1 \dots d}  
\end{equation}
where $\phi$ represents the function for performing object detection; $\mathbf{Y}$ is composed of classification and regression components, respectively; and $d$ represents the number of objects to detect in a sample. $\phi$ can take the form of conventional fusion algorithms, a machine learning model, or an ensemble of machine learning models.
Classification refers to the identification of each detected object's class. The classification target for each object can be defined as:
\begin{equation}
    \mathbf{Y}_{class}^i \in \{1, 2, 3, \dots , k\}   
\end{equation}
where $k$ represents the number of classes considered in the problem. These indices represent a pre-defined mapping to object classes (\textit{e.g.}, $1$:\textit{car}, $2$:\textit{van}, $3$:\textit{truck}, and so forth). 
Regression refers to the estimation of an object's location within the sample. These targets can be represented by:
\begin{equation}
    \mathbf{Y}_{reg}^i = [\mu_1 , \nu_1 , \mu_2, \nu_2 ] \in \mathbb{R}^2  
\end{equation}
where $\mu$ and $\nu$ denote the object's 2D bounding box coordinates in reference to a generic coordinate frame. \footnote{This could be represented in 3D as well, but for the sake of this paper we focus on 2D object detection.} 

The measurements from $s$ sensors can be fused by a variety of means to improve detection results. An early fusion approach involves fusing the raw sensor measurements before passing them to $\phi$:
\begin{equation}
    \mathbf{Y}  = \phi(\psi(\mathbf{X}_1, \mathbf{X}_2, \dots , \mathbf{X}_s ))   
\end{equation}
with $\psi$ representing the function used to fuse the measurements. 
In the case of late fusion, $\hat{\phi}$ represents a function fusing the separate output detections:
\begin{equation}
    \mathbf{Y}_1,\, \mathbf{Y}_2,\, \dots, \mathbf{Y}_s = \phi_1(\mathbf{X}_1),\,  \phi_2(\mathbf{X}_2),\, \dots , \, \phi_s(\mathbf{X}_s)
\end{equation}
\begin{equation}
    \mathbf{Y} = \hat{\phi}(\mathbf{Y}_1, \mathbf{Y}_2, \dots , \mathbf{Y}_s)
\end{equation}

The context of scenes in AV driving can vary dramatically: from different lighting conditions, to different road types and locations, to weather conditions that can severely degrade specific sensors. This variance calls for the use of an adaptive $\phi$ that is not only determined by a set of static scene conditions, but is instead learned within the model. 
In this case, $\phi$ represents an ensemble of object detection models, and $\phi^*$ represents the expected best subset of models in the ensemble for a given input $\mathbf{X}$.
We denote the contextual information of a scene (either learned and modeled from the inputs or provided externally) as $\Omega$. We then can define the subsequent equations:
\begin{equation}
    \Omega = \pi(\mathbf{X}) ,\quad \phi^* = \rho(\Omega) ,\quad \mathbf{Y} = \phi^*(\mathbf{X})
\end{equation}
where $\pi$ represents a context identification model, and $\rho$ represents the mechanism for selecting $\phi^*$ given the identified context $\Omega$.
The goal of $\pi$ and $\rho$ is to select the optimal subset of branch models $\phi^*$ for the inferred context $\Omega$ to maximize object detection performance for a given $\mathbf{X}$. We posit that this general problem formulation can be extended to other sensor fusion problems in CPS.



\section{Methodology}
\label{methodology}
\begin{figure}
    \centering
    \includegraphics[clip, trim=10 100 490 15, width=\linewidth]{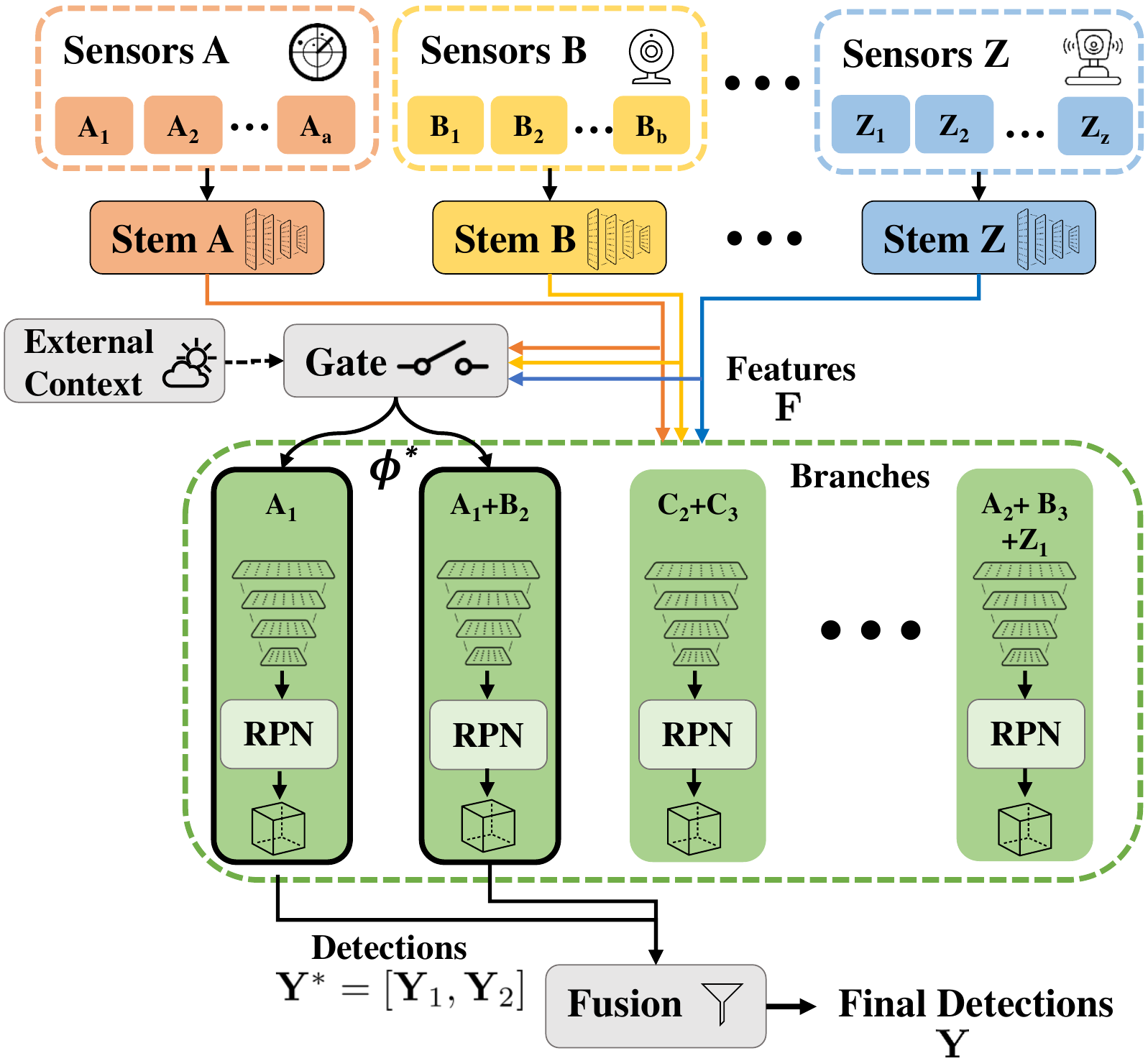}
    \caption{Our Proposed \textsc{HydraFusion} Architecture.}
    \label{fig:archi}
    \vspace{-2mm}
\end{figure}
The model architecture for our proposed approach, \textsc{HydraFusion} is shown in Figure \ref{fig:archi}. Algorithm \ref{alg:hydrafusion} describes how our architecture processes input data from different modalities to produce the desired targets. 
First, sensor data from each modality is processed by a modality-specific CNN (denoted as ``stem'') to produce an initial set of features $\mathbf{F}$. Next, these features are used by the gating module (containing $\pi$ and $\rho$) to identify the context $\Omega$ and select which subset of branches $\phi^*$ should be executed for this context.
Each branch is a deep-learning model capable of converting the features extracted from a subset of sensors $\mathbf{F}^*$ to produce a set of outputs for a specific task (\textit{e.g.}, object detection). After the selected subset of branches is executed, the branches pass their outputs $\mathbf{Y}^*$ to the fusion block, which fuses them to generate the final object detections $\mathbf{Y}$. Next, we discuss the details of each component in our proposed architecture.

\begin{algorithm}
\caption{\textsc{HydraFusion} Algorithm}
\DontPrintSemicolon
\label{alg:hydrafusion}
\KwIn{Sensor measurements $\mathbf{X} = \{X_{A_1}, X_{A_2}, X_{B_1},  \dots, X_{Z_m}\}$}
\KwOut{Object Detections $\mathbf{Y}$}

$\mathbf{F} \gets [[],[],[],...]$ \tcp*{initialize feature vector}

\For {s in sensor\_types}{ 
$\mathbf{S} \gets \mathbf{X}[s]$ \tcp*{get data by modality}
\For{m in S}{
$F[s][m] \gets  stem_s(m)$ \tcp*{extract features}
}
}
$\Omega \gets \pi(\mathbf{F})$ \tcp*{identify context}
$\phi^* \gets \rho(\Omega)$ \tcp*{select Top-$k$ branches to run}
$\mathbf{Y}^* \gets []$\;
\For {branch in $\phi^*$}{
$\mathbf{Y}^*[branch] \gets branch(\mathbf{F^*})$\tcp*{pass subset of $\mathbf{F}$}
}

$\mathbf{Y} \gets fusion\_block(\mathbf{Y}^*)$ \tcp*{fuse branch detections}
\end{algorithm}

\subsection{Input Processing and Stems}
As shown in Figure \ref{fig:archi}, \textsc{HydraFusion} accepts any number of sensors and sensing modalities as input.  
Each stem is implemented as a CNN, which generates an initial set of spatial features for each sensor.
We use a shared stem block for processing all the sensors for a given sensor modality. Thus, we will have three stems if our implementation uses camera, radar, and lidar sensors. After the input from each sensor for a given modality is passed through the stem, the gate module uses the resulting features to identify the context and select which branches to execute. Then, the selected branches use the stem output features as inputs to generate their predicted object detections. 

\subsection{Context Identification and Gating Module}
\label{subsec:gating}

\begin{figure}[htb]
    \centering
    \includegraphics[clip, trim=20 652 1560 30, width=\linewidth]{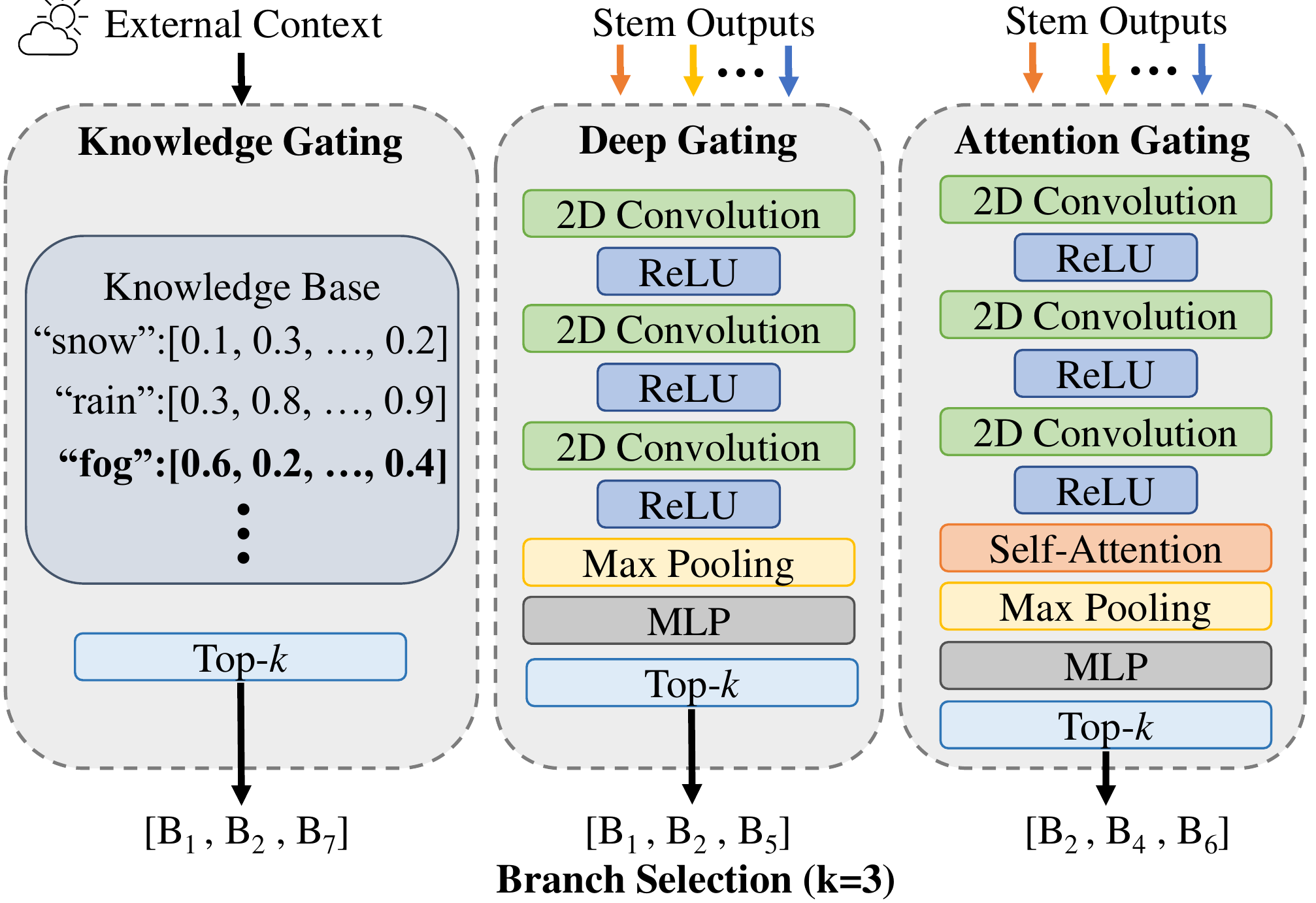}
    \caption{Gating Model Architectures.}
    \label{fig:gating_architectures}
    \vspace{-4mm}
\end{figure}

Context identification is important for selecting the appropriate subset of branches to maximize performance in a given context. We propose several different gating algorithms for this task. 
The goal of the gate module is to rank the branches based on their expected performance for the input set of stem features. Next, the top-$k$ branches (where $k$ is configurable) are selected for execution and fusion to maximize object detection performance. The architectures of our three gating models are shown in Figure \ref{fig:gating_architectures}.

\paragraph{Rigid Knowledge-Based Gating}
Since there exists some domain knowledge as to how each context will affect each sensing modality, we can implement \textit{Knowledge Gating}, where this domain knowledge is used to statically encode the subset of branches to execute for a given context. This assumes the set of possible contexts is finite, and the current context can be identified via external sources. For example, weather information, time of day, and map data can all be used to define the current context. In our approach, we define the set of fixed contexts based on metadata from the RADIATE dataset \cite{sheeny2020radiate} describing the type of driving data in each sequence. Thus, our set of fixed contexts is: \{\textit{city, motorway, junction, rural, snow, fog,} and \textit{night}\}. We leverage domain knowledge from Table \ref{tab:mot_tab} as well as from the RADIATE paper to rank the relative performance of each sensor in each fixed context. Then, at run-time, the external context information (\textit{e.g.}, data from a navigation/weather system) is used to identify the current context. The top-$k$ ranked branches for that context are selected to be executed and fused. The limitation of this gating strategy is that it requires a \textit{fixed} context definition, potentially limiting performance in cases where contexts are less rigidly defined. With our other gating strategies, we define the context as a \textit{continuous} feature space to enable the modeling of more complex contexts.

\paragraph{Learned Dynamic Deep Gating}
In \textit{Deep Gating}, we implement a CNN followed by a multi-layer perceptron (MLP) to model the relationship between the features output from the stems and rank the branches based on their expected performance for this feature set. In this gating method, the context can be viewed as a continuous feature space defined by the stem outputs.

\paragraph{Attention-Based Dynamic Gating}
In some contexts, certain regions of the feature map may be more informative than others about the scene's context and, consequently, the branch-wise performance. We implement an attention-based gating strategy, denoted as \textit{Attention Gating}, that infers an attention map over the stem features to evaluate this hypothesis. This attention map is used with CNN and MLP layers to model the relationship between branch performance and stem features. We use the visual attention layer proposed in \cite{zhang2019self} in our implementation.

\paragraph{Optimal Loss-Based Gating}
To serve as a performance target for our gating approaches, we implement a so-called "optimal" gating strategy where, for a given input, the branch ranking output by the gate module is equal to the inverse of the aggregated branch loss for the detections output by each branch. Since the actual branch loss is used to inform the gate \textit{a posteriori}, this strategy is not feasible for real-world deployment.  However, it gives the theoretical best-case performance of a gating strategy that can perfectly rank the branches based on their losses for a given input. We denote this gating method \textit{Optimal Gating}.

\subsection{Branches}
The branches of the proposed framework are designed to be specific to different sensor fusion combinations. These pairings can enforce early fusion in the model by combining the stem features of heterogeneous sensor inputs (\textit{e.g.}, radar and lidar) before performing object detection. Furthermore, some branches use singular sensor inputs (\textit{e.g.}, radar) that the gating module may choose in scenarios where other sensors (\textit{e.g.}, camera and lidar) have poor performance due to situational factors (\textit{e.g.}, weather or obstruction). Each branch is equipped with a Region Proposal Network (RPN) \cite{ren2015faster} that uses anchor generation techniques to predict detections across a feature map. These predictions are then fed through a region-of-interest layer that generates the following outputs for each detection: \textit{bounding box coordinates} $[\mu_1, \nu_1, \mu_2, \nu_2]$ --- expressed in the native coordinate frame, \textit{scores} $[0-1]$ --- confidence level of the detected object, and \textit{labels} \{1, 2, 3, \dots , k\} --- the assigned classification of the object. The outputs from each branch are passed to the fusion block to generate the final set of fused detections.


\subsection{Fusion Block}
The function of the fusion block in our approach is synonymous with the concept of late fusion. In \textsc{HydraFusion}, we use the following fusion algorithms to fuse the detections output by all of the active branches of the model. 




\paragraph{Non-Maximum Suppression (NMS)}
This algorithm calculates the intersection over union (IoU) of corresponding bounding box estimations, and based on their confidence scores, selects which box estimates to keep. The equation for calculating the IoU (sometimes referred to as the Jaccard index) between two sets, $A$ and $B$, is given by:
\begin{equation}
    IoU(A,B) = \frac{|A \cap B|}{|A \cup B|} , 
\end{equation}
where $\cap$ represents the intersection, and $\cup$ represents the union. In our application, the sets are the rectangular bounding box predictions. By iteratively comparing bounding box predictions and returning a match if the IoU is above a defined threshold, only the box with the highest confidence score is kept among each set. 
\paragraph{Soft-NMS}
A further refinement of NMS, proposed in \cite{bodla2017soft}, which lowers confidence scores using a Gaussian weighting function defined by $\sigma$, if the boxes are above a threshold IoU value. Unlike NMS, Soft-NMS does not completely remove box estimates, which can result in more false positives. 
\paragraph{Weighted Box Fusion (WBF)}
This approach, proposed in \cite{solovyev2021weighted}, clusters the bounding box predictions into distinct lists by iterating over the boxes and calculating IoUs with respect to thresholds. From each cluster, the fused bounding box predictions, $[\mathbf{f}_\mu , \mathbf{f}_\nu]$, are computed as weighted sums of each detection and its confidence score:  
\begin{equation}
    f_{{\mu}_j} = \frac{\sum_{i=1}^{n} C_i \cdot \mu_{i,j}}{\sum_{i=1}^{n} C_i} ,\quad f_{{\nu}_j} = \frac{\sum_{i=1}^{n} C_i \cdot \nu_{i,j}}{\sum_{i=1}^{n} C_i} 
\end{equation}
where $j \in \{1,2\}$; $\mu_{i,j}$ and $\nu_{i,j}$ are the corresponding locations of the bounding box points; and $C_i$ is the confidence score for the $i$th box. WBF also has a skip-box threshold that defines which boxes to exclude if they are below a certain confidence score. Furthermore, each branch can be assigned varying weights that can be tuned within the overall model or application being used. In our experiments, covered in the next section, the tunable threshold parameters across fusion methods were found to have insignificant effects on the results within reason. 

\section{Experiments}
\label{exp}
In this section, we discuss our experiments. In Section \ref{datasets} we elaborate on the dataset used to conduct the experiments. Sections \ref{subsec:model_impl} and \ref{training} detail our model implementation process and training procedures. In Section \ref{results} we present our experimental results. Finally, in Section \ref{discussion} we discuss the practicality of our approach and future work.

\subsection{Dataset}
\label{datasets}
The RADIATE dataset \cite{sheeny2020radiate} contains annotated data from a Navtech CTS350-X radar, a Velodyne HDL-32e LiDAR, and a ZED stereo camera. With this dataset, we trained and evaluated our models on object detection using supervised learning. The RADIATE dataset contains data for various driving contexts, including urban driving, snow, rain, fog, night, and motorway driving. In some cases, several sensors are visually obstructed by fog, rain, or snow. The dataset contains the following annotated object classes: \{\textit{car, van, truck, bus, motorbike, bicycle, pedestrian, group of pedestrians}\}. 
This dataset provides a challenging benchmark on which the robustness of object detection models can be evaluated for a range of driving contexts. They additionally present object detection results using radar in varying weather conditions.
Since \cite{sheeny2020radiate} uses a different problem formulation, model size, and metrics, its results are not directly comparable to ours; however, our results for radar-only are representative of the model evaluated in their work. 
Please refer to the dataset for further details \cite{sheeny2020radiate}. We used a 70:30 train-test split for training and evaluating our models.


\subsection{Model Implementation}
\label{subsec:model_impl}
\subsubsection{Model Specification}
To evaluate \textsc{HydraFusion} in comparison to the baseline fusion approaches, we implemented each stem and branch as a Faster R-CNN \cite{ren2015faster} model with a ResNet-18 \cite{he2016deep} backbone. We split the ResNet-18 models at the first block and use it as the stem for each modality. Then, the remaining ResNet-18 layers and the RPN of Faster R-RCNN are used in each branch. 

With four sensors (two cameras, lidar, and radar), the total number of possible unique branches is $2^4 - 1 = 15$. However, the training and space complexity of a 15-branch model may be much larger without providing noticeable improvements in precision. Thus, we use domain knowledge to identify the best branches for the application by picking branches that can cover the limitations of other branches in difficult contexts. 
Thus, our \textsc{HydraFusion} implementation contains four single-sensor branches and three early fusion branches, for a total of seven branches. The single-sensor branches are: \textit{Left Camera, Right Camera, Lidar}, and \textit{Radar}; the early fusion branches are \textit{L/R Cameras, Lidar+Radar}, and \textit{L/R Cameras + Lidar}.
For single-sensor branches, the stem features for the sensor are used as the input for the branch.
For branches with early fusion, we concatenate the stem features for each sensor to be fused across the channel dimension. Then, we use a 2D convolution layer to fuse this concatenated output before passing the result to the remaining ResNet-18 layers in the branch. 

Regarding the fusion block, the three fusion algorithms we implemented used the following thresholds during the experiments: IoU threshold = 0.4, skip-box threshold = 0.01, $\sigma = 0.5$. Due to computation constraints, we only evaluated ResNet-18 in this work; however, this architecture can be directly used with larger image-processing models (\textit{e.g.}, ResNet-34/50/152, DenseNet-169, VGG-16) by simply changing the image processing backbone and picking a different split-point to divide it between the stems and the branches.

\subsubsection{Gating Module Specification and Training}
We implemented deep convolutional networks for the Deep and Attention Gating methods. As shown in Figure \ref{fig:gating_architectures}, the Deep Gating model is implemented as a 3-layer CNN with an MLP layer to map the CNN output to seven output channels, corresponding to the branch ranking for the seven branches. The Attention Gating method differs in that a self-attention layer is added after the CNN but before the max pooling and MLP layers.
We trained the Deep and Attention Gating methods to estimate the aggregated loss of each branch for a given input using regression with mean absolute error as the loss function. The top-$k$ lowest loss branches predicted by the gate were selected for fusion. To prevent the gate model training process from affecting the training process of the \textsc{HydraFusion} model, we trained and evaluated the gating modules separately using the stem outputs and branch losses of a fully trained \textsc{HydraFusion} model as the inputs and targets for the gate. After training, the gate model can be re-introduced into the \textsc{HydraFusion} model for deployment. 

As mentioned in Section \ref{subsec:gating}, the Knowledge Gating approach uses external context and domain knowledge to inform the branch ranking. During inference, we query the knowledge base using the external context for each input and return the branch rankings defined for that context. 
For the Optimal Gating method, we take the loss between the ground-truth boxes and the branch outputs for each branch and use this information to rank the branches --- the branches with the lowest aggregated loss are ranked the highest.

\subsubsection{Perspective Mapping}
Since the RADIATE dataset contains data from both forward-facing (stereo cameras) and birds-eye view (radar and lidar) perspectives, we used a transformation matrix to transform the predicted bounding boxes from the birds-eye view (BEV) sensors to the forward-facing perspective (FWD). This enabled us to fuse the detections from both perspectives in the fusion block. 
To allow a fair assessment in our analysis across the different sensor modalities, we chose the cameras' field of view to be the fused reference frame as it was the limiting factor since it covers the least area. This prevents the objects detected by the lidar and radar branches from dominating the model when objects are detected outside the cameras' field of view. The transformations from the various other sensors to the reference frame are detailed further in Appendix \ref{trans}. We postulate that our approach could be directly applied in scenarios with full 360-degree camera coverage without loss of generality.

\subsection{Training and Scoring Metrics}
\label{training}
We built, trained, and evaluated each model in PyTorch using a batch size of 1 with a learning rate of 5e-3 for training the stem/branch models and 5e-5 for training the gate models. We computed classification and box regression loss using the multi-task loss function defined and used in Faster R-CNN \cite{ren2015faster}. Please reference their work for more details on the loss calculation.


To score the models on object detection, we used the mean average precision (mAP) score, which is widely utilized as the primary metric for benchmarking object detection models \cite{ren2015faster, everingham2010pascal}. We compute the mAP for boxes with an intersection-over-union (IoU) $\geq$0.5, which aligns with the PASCAL Visual Object Classes (VOC) Challenge \cite{everingham2010pascal}. 
Precision ($P$) and recall ($R$) for each class in the dataset are defined as: 
\begin{equation}
    P = TP/(TP+FP), \; R = TP/(TP+FN)
\end{equation}
where $TP$, $FP$, and $FN$ represent the number of true positive, false positive, and false negative classifications, respectively, by the model at a set confidence threshold. 
Average precision (AP) is a measure of the area under the precision-recall curve and is calculated as follows:
\begin{equation}
\text{AP} = \sum_n (R_n - R_{n-1}) P_n
\end{equation}
where $R_n$ and $P_n$ correspond to the recall and precision at threshold $n$ on the precision-recall curve. We calculate the mAP as the mean of the AP across all object classes where every object instance is weighted equally. 

\subsection{Results}
\label{results}
Here we present our experimental results for object detection, evaluate our proposed gating methods, and benchmark our model's performance on industry-standard AV hardware.

\subsubsection{Object Detection Results}
\label{object_detection}
In Table \ref{tab:map_scores}, we show the mAP achieved by different model configurations on the dataset. Results are shown for (i) individual sensors, (ii) early fusion between sensors, (iii) late fusion between sensor-specific branches, and (iv) our proposed \textsc{HydraFusion} approach. For the results in sets (i) and (ii), the mAP is calculated from a single ResNet-18 FasterRCNN model taking the stated sensor data as input. The late fusion results are computed by processing each sensor modality separately through a ResNet-18 FasterRCNN model and fusing the outputs of each model using one of the three fusion algorithms (WBF, NMS, or Soft-NMS), with the best performing result shown in the table. All-Branches (Early + Late) is the result from running all of the branches in \textsc{HydraFusion} and fusing the results using the fusion algorithms. Set (iv) shows the results for our selective sensor fusion approach using the Attention Gating method to select the Top-3 branches for each input. 

\begin{table}[htb]
    \centering
    \begin{tabular}{p{75pt} l c}
    \hline
        Fusion Method & Model & mAP \% \\\hline
        \multirow{3}{*}{(i) No Fusion} 
        & Single Camera & 65.33\\
        & Radar & 69.42\\
        & Lidar & 61.86\\\hline
        \multirow{3}{*}{(ii) Early Fusion} 
        & L/R Cameras& 65.33\\
        & Radar + Lidar & 71.63\\
        & Camera + Lidar & 65.99\\\hline
        \multirow{5}{*}{(iii) Late Fusion} 
        & L/R Cameras & 65.71\\
        & Radar + Lidar & 65.33\\
        & L/R Cameras + Lidar & 66.20\\
        & Radar + Lidar + L/R Cameras & 71.16\\
        & All-Branches (Early + Late) & 65.47 \\\hline
        \multirow{3}{75pt}{\textbf{(iv) \textsc{HydraFusion} (Ours)}} 
        & \textbf{Top-3 Branches w/ WBF} & \textbf{74.54} \\
        & \textbf{Top-3 Branches w/ NMS} & \textbf{78.51} \\
        & \textbf{Top-3 Branches w/ Soft-NMS} & \textbf{81.31} \\\hline
    \end{tabular}
    \caption{Object detection mAP scores on the RADIATE dataset for: (i) single sensors, (ii) early fusion, (iii) late fusion, and (iv)   \textsc{HydraFusion} (ours) with Attention Gating.}
    \label{tab:map_scores}
    \vspace{-5mm}
\end{table}

\begin{figure}[htb]
    \centering
    \includegraphics[clip, trim=20 300 430 50, width=\linewidth]{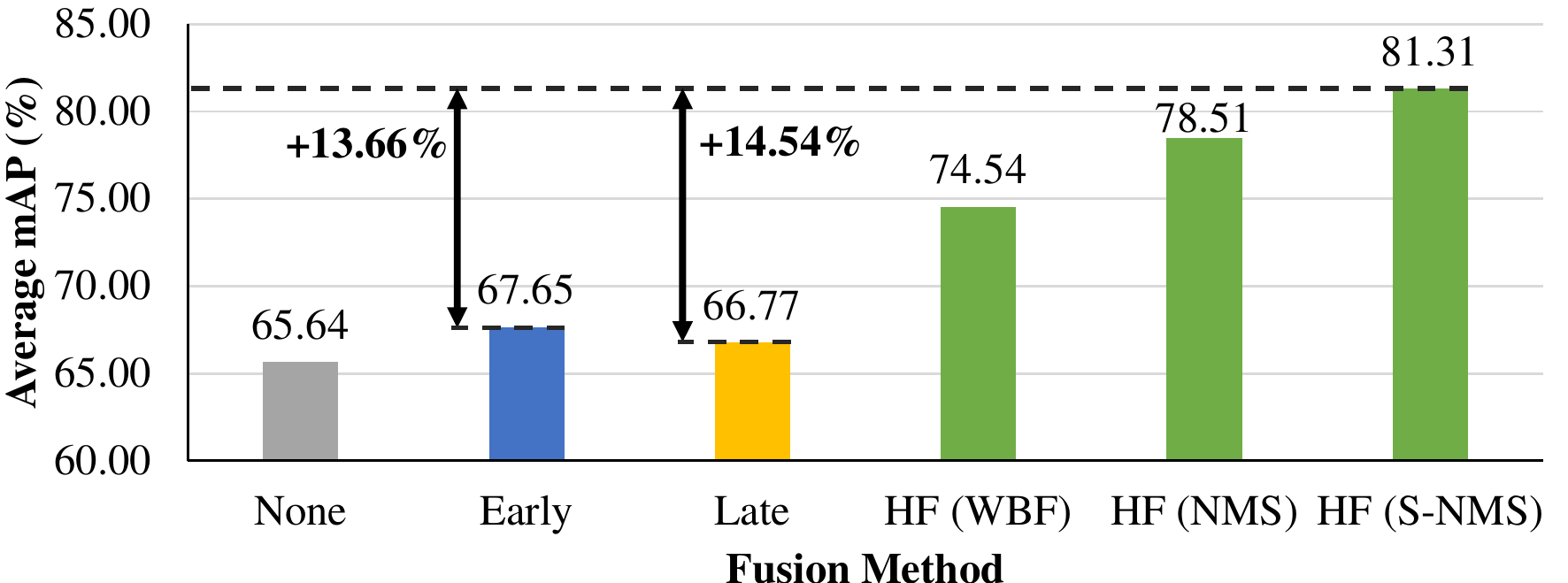}
    \caption{Average mAP for each fusion method compared against \textsc{HydraFusion} (HF).}
    \label{fig:avg_map}
    \vspace{-2mm}
\end{figure}

Interestingly, All-Branches performs \textit{worse} than all the results in (iv), supporting our hypothesis that using less sensor data can improve robustness. The tradeoffs between early and late fusion approaches are also shown. Early fusion can perform better with fewer sensors if the sensors provide good quality data (Radar + Lidar). In comparison, late fusion is more robust to bad data but requires more sensors to achieve good performance (Radar + Lidar + L/R Cameras). The table also shows the benefits of fusion compared to single-sensor approaches, as all fusion variants outperform (i) in at least one configuration. 
Figure \ref{fig:avg_map} shows the average mAP for each fusion method.
As shown, \textsc{HydraFusion} significantly outperforms both early and late fusion approaches on average (by 13.66\% and 14.54\%, respectively), achieving a peak mAP of \textbf{81.31\%}.
Overall, the results support our hypothesis that a context-aware selective sensor fusion architecture is significantly more robust and accurate than existing fusion methods.

\subsubsection{Gating Method Evaluation}
\label{gating_evaluation}

\begin{table}[htb]
\centering
\begin{tabular}{p{110pt} c c c c }
\hline
\multirow{2}{*}{Gate Model (Fusion Alg.)} & \multicolumn{4}{c}{mAP \%}\\\cline{2-5} 
& $k=1$ & $k=3$ & $k=5$ & $k=All$\\\hline
Knowledge Gating (NMS) &77.59 &76.37 &76.53 &75.81\\
Deep Gating (NMS) &67.43 &78.14 &73.31 &75.81\\
\textbf{Attn. Gating (Soft-NMS)} &67.27 & \textbf{81.31} &69.88 &65.71\\\hline
Optimal Gating (Soft-NMS) &73.03 & \textbf{81.57} &72.93 &65.71\\\hline
\end{tabular}
\caption{Gating evaluation for different $k$. The highest mAP indicates which gate configuration is best for real-world deployment of \textsc{HydraFusion}.}
\label{tab:gating_eval}
\vspace{-3mm}
\end{table}

Next, we evaluate our proposed gating strategies.
To evaluate the impact of different subset sizes $k$ on each of our proposed gating methods, we computed the mAP after fusion for $k \in \{1, 3, 5, All (7)\}$ with WBF, NMS, and Soft-NMS fusion.
The results for the best performing fusion algorithm for each gating approach are shown in Table \ref{tab:gating_eval}.
Optimal Gating represents the theoretical best performance if the $k$ lowest-loss branches are selected for each input. 

As shown, Attention Gating using Soft-NMS achieves the best mAP for 3-branch fusion, with a score of \textbf{81.31\%} (only 0.26\% less than Optimal Gating). This likely results from its capability to identify the regions in the input that are most relevant to the output. Deep Gating was the second-best approach with a mAP of 78.14\% for 3-branch fusion as it was still able to identify the context well using the stem features. 

Interestingly, Knowledge Gating performed best for $k=1$, likely because the domain knowledge was sufficient to determine the best modality for each context. However, Knowledge Gating did not achieve as high of a mAP score as Deep and Attention Gating for any $k$, meaning that its performance across contexts is generally worse. Besides, in real-world deployments, $k=1$ would be insufficiently robust to sensor obstruction or failures, so $k=1$ performance is less relevant to real-world use cases than $k \in \{3,5,All\}$ performance. For our application, $k=5$ and $k=All$ did not perform as well as $k=3$. Overall, the results in Table \ref{tab:gating_eval} show that Attention Gating with 3-branches results in the highest object detection mAP score (4.94\% higher than Knowledge Gating) and is thus the best configuration to use on an actual vehicle.

\subsubsection{Hardware Energy and Latency Evaluation}
\label{hardware_evaluation}
\begin{figure}[ht]
    \centering
    \includegraphics[width=0.75\linewidth]{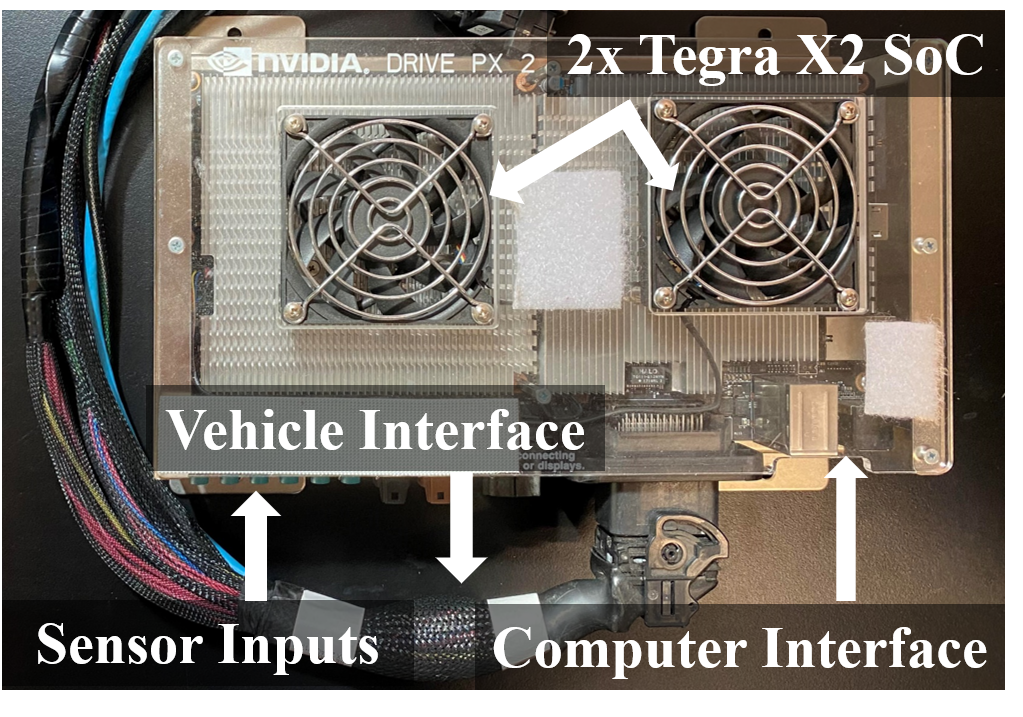}
    \caption{Nvidia Drive PX2 Testbed.}
    \label{fig:hardware}
    \vspace{-5mm}
\end{figure}

\begin{table}[htb]
    \centering
    \begin{tabular}{p{28pt} p{96pt} p{22pt} p{24pt} p{24pt}}
    \hline
        Fusion Method & Configuration & Energy (J) & Latency (ms) & Memory (MB)\\\hline
        \multirow{2}{26pt}{None} 
        & Radar or Lidar & 0.954 & 21.85 & 769\\
        &Single Cam. & 0.945 & 21.57 & 767\\\hline
        \multirow{3}{26pt}{Early Fusion}& 
        L/R Cam. & 1.192 & 27.36 & 768\\
        &L/R Cam. + Lidar & 1.379 & 31.36 & 694\\
        &L/R Cam. + Lidar + Radar & 1.615 & 36.86 & 750\\\hline
        \multirow{3}{26pt}{Late Fusion}&
        L/R Cam. & 1.959 & 43.99 & 923\\
        &L/R Cam. + Lidar & 2.878 & 64.09 & 1087\\
        &L/R Cam. + Lidar + Radar & 3.769 & 84.32 & 1239\\\hline
\multirow{4}{26pt}{\textsc{Hydra-}\textsc{Fusion}} 
        & 3-Branch (Deep Gating) & 3.317 & 73.84 & 1271\\
        & \textbf{3-Branch (Attn. Gating)} & \textbf{3.284} & \textbf{73.02} & \textbf{1080}\\
        & 5-Branch (Deep Gating) & 5.008 & 110.58 & 1390\\
        & 5-Branch (Attn. Gating) & 4.897 & 107.28 & 1390\\
        \hline
    \end{tabular}
    \caption{Hardware evaluation on the Nvidia Drive PX2. Reported numbers are for processing one input through the model.}
    \label{tab:hardware}
    \vspace{-4mm}
\end{table}

To demonstrate that our approach is practical for real-world deployment, we analyze the energy consumption, latency, and memory usage of our model on an industry-standard AV hardware platform, the Nvidia Drive PX2, shown in Figure \ref{fig:hardware}. To perform hardware analysis, we compiled each model specification using TensorRT and used built-in tools to measure its end-to-end latency and memory usage. Then, we multiply this value by the power consumption of the system measured via an external power meter to obtain the energy consumption.

In Table \ref{tab:hardware}, we show the results for running different model variations including single sensor models, early fusion models, late fusion models, and our \textsc{HydraFusion} methodology. 
The \textsc{HydraFusion} 3-branch results shown are for the worst-case energy and latency scenario where all three branches selected by the gate are early-fusion branches. Similarly, the \textsc{HydraFusion} 5-Branch result is with three early-fusion branches and two single-sensor branches selected. The \textsc{HydraFusion} results are shown with Deep Gating and Attention Gating modules.

As expected, the single-sensor and early fusion methods are the least demanding on hardware since they only use a single ResNet-18 Faster R-CNN model; however, they also achieve lower mAP scores overall, as shown in Table \ref{tab:map_scores}.
The results show that the \textsc{HydraFusion} 3-Branch configurations have energy consumption, latency, and memory usage that is comparable to 3-sensor and 4-sensor late fusion models. This result means 3-branch \textsc{HydraFusion} can reasonably be used in cyber-physical systems where late fusion approaches are currently deployed. Since 3-branch \textsc{HydraFusion} achieves significantly higher mAP than both early and late fusion methods, it presents clear benefits over state-of-the-art methods. 
The 5-branch \textsc{HydraFusion} was slower and less energy efficient than 3-branch and also achieved a lower mAP score (as shown in Table \ref{tab:gating_eval}), so 3-branch would be preferred for real-world implementation.
For both 3-branch and 5-branch \textsc{HydraFusion}, Attention Gating was slightly more efficient than Deep Gating, likely because TensorRT better optimized its architecture.

\subsection{Discussion}
\label{discussion}
\subsubsection{Practicality}
As mentioned in Section \ref{hardware_evaluation}, the energy, latency, and memory usage of \textsc{HydraFusion} on the industry-standard Nvidia Drive PX2 is comparable to that of late fusion --- meaning that \textsc{HydraFusion} can be used in any CPS where late fusion is currently in use. 
Thus, to implement \textsc{HydraFusion} in a real AV, the trained model and hardware can be installed in the vehicle and integrated with the perception module of the existing modular AV software stack. 
Although we evaluated one hardware platform with four sensors in our experiments, our approach is hardware- and sensor-agnostic. It can be used with any hardware platform and sensor configuration by using the corresponding model compilation tools and aligning the sensor data to the model's input. Additionally, our approach can be applied to a wide range of CPS problems besides object detection. Any CPS application using sensor fusion can potentially benefit from our context-aware selective sensor fusion approach. 
The model size and memory requirements will increase proportionally with more sensors due to the increased number of branches. However, they will likely still be comparable to late fusion with the same number of sensors. The branches must also be defined using domain knowledge for the new task; for example, sensors that cover the same FOV or complement each other can be combined to form early fusion branches.

\subsubsection{Limitations and Future Work}
We statically defined the set of branches used in \textsc{HydraFusion} for AV object detection using domain knowledge. Thus, our approach does not enable selecting between every possible set of sensor combinations for each branch. Doing so would not be computationally feasible as the space complexity would be $O(2^n)$, and the training time would increase similarly. Thus, our approach currently requires domain knowledge to identify the subset of branches that provide the most coverage across scenarios without exceeding model complexity or size requirements. Future research could explore automated techniques for selecting the optimal set of branches to use in the model.
In this paper, we focused on the problem of object detection for AVs; however, our approach can be directly applied to a wide range of multi-modal CPS and internet-of-things (IoT) problems. Different backbone models or fusion methods can be used to enable \textsc{HydraFusion} to model new tasks, such as tracking, localization, and control. 
We also believe that improved gating strategies with temporal modeling components could provide avenues for improving context identification, task performance, and resource utilization.
It would also be prudent to evaluate the difference in safety between our approach and existing methods, especially in challenging driving conditions.

\section{Conclusion}
\label{sec:conclusion}
In this paper, we present \textsc{HydraFusion} --- a sensor fusion framework that can selectively fuse sensor inputs in a context-aware manner. We validate our approach through theoretical, qualitative, and quantitative analysis on the task of object detection performed by AV perception systems on a challenging and diverse real-world dataset. 
On average, our selective sensor-fusion approach achieved a mAP score \textbf{13.66\%} and \textbf{14.54\%} higher than early fusion and late fusion approaches, respectively, supporting our hypothesis that a context-aware selective sensor fusion approach improves robustness.
Additionally, we proposed and evaluated several gating models to perform context identification and branch selection, finding that an attention-based deep learning gate model was \textbf{4.94\%} more effective than static selection methods. Lastly, we evaluated our proposed approach on industry-standard AV hardware, showing that our approach had comparable energy consumption, latency, and memory usage to existing fusion architectures. Ultimately, \textsc{HydraFusion} offers a novel sensor fusion approach for multi-modal CPS that can not only improve performance but also help support safer autonomous driving.




\bibliographystyle{ACM-Reference-Format}
\bibliography{bibliography}

\appendix

\section{Sensor Coordinate Frame Transformations}
\label{trans}
The transformations to convert the detections from one sensing modality to the reference frame occur in three main steps, which are detailed for the radar sensor as follows. Firstly, a point in the radar pixel coordinates, $[u,v]^r$, is transformed into the radar Cartesian frame:
\begin{equation}
    [x,y]^r = \gamma \cdot ([u,-v]^r - [w/2, -h/2] ) ,
\end{equation}
where $\gamma$ is the radar resolution expressed as meters/pixel, $w$ is the width of the radar image in pixels, and $h$ is the height of the radar image in pixels. An additional step to add a height, $z^r$, is computed by mapping the object's classification to a defined set of average class heights. Secondly, this Cartesian representation in the radar coordinate frame must be expressed in the chosen reference frame --- the camera Cartesian frame. This is accomplished by subsequent the translation and rotation of coordinate frames via the following:
\begin{equation}
    [x,y,z]^c = R^c_r \cdot ([x,y,z]^r + T^r ) , 
\end{equation}
where the superscript $c$ indicates the world camera frame, $R^c_r$ is the $3\times3$ rotation matrix from the radar to the camera frame, and $T^r$ is the $1\times3$ translation vector between the radar and camera. Thirdly, the intrinsic parameters of the camera are used to convert the points from the Cartesian camera frame to the pixels in the image frame:
\begin{equation}
    s\begin{bmatrix}u \\ v \\ 1 \end{bmatrix}^c = P \cdot \begin{bmatrix}x \\ y \\ z \end{bmatrix}^c , 
\end{equation}
\begin{equation}
    P = \begin{bmatrix}
            f_x & 0 & c_x\\
            0 & f_y & c_y\\
            0 & 0 & 1
        \end{bmatrix} 
\end{equation}
where $s$ is an arbitrary scaling factor and $P$ is the camera intrinsic projection matrix, which is constructed during calibration of the camera using camera focal length parameters $(f_x, f_y)$ along with the principal point, or optical center of the camera $(c_x,c_y)$. Using two stereo cameras provides the ability to derive depth information from the image to complete the necessary transformation presented above. The same procedures are repeated for the other sensors, with the respective adjustments to the translation and rotation vectors as needed. We note that this transformation process introduces additional uncertainties into the fusion as factors such as road elevation can alter the result in certain scenarios.


\section{Additional Results}
\subsection{Extended Object Detection Results}
Table \ref{tab:map_scores2} expands on the late fusion results from Table \ref{tab:map_scores} and shows a comparison between different late fusion algorithms for the chosen sensor combinations. Here, we examine results using WBF, NMS, and Soft-NMS across the different models. Only the fourth model, Radar + Lidar + L/R Cameras, has a noticeable improvement using WBF, while the other variations remained within similar score ranges. We attribute the closeness of these results to the three chosen fusion algorithms having similar statistical techniques embedded within them.

\begin{table}[htbp]
\centering
\begin{tabular}{p{120pt} c c c }
\hline
\multirow{1}{*}{Model} & \multicolumn{3}{c}{mAP \%}\\\cline{2-4} 
& WBF & NMS & Soft-NMS \\\hline
L/R Cameras &65.71 &65.71  &65.71  \\
Radar + Lidar &65.33 &65.33 &65.33 \\
L/R Cameras + Lidar &66.06 &\textbf{66.20} &66.18 \\
Radar + Lidar + L/R Cameras &\textbf{71.16} &67.11 &65.42 \\
All-Branches &64.85 &63.64 &\textbf{65.47} \\\hline
\end{tabular}
    \caption{Object detection mAP scores on the RADIATE dataset for different late fusion algorithms: (i) WBF, (ii) NMS, (iii) Soft-NMS.}
    \label{tab:map_scores2}
     \vspace{-5mm}
\end{table}

\subsection{Extended Gating Results}

\begin{table}[!h]
\centering
\begin{tabular}{p{68pt} p{38pt} c c c c }
\hline
\multirow{2}{*}{Gate Model} & Fusion Alg. & \multicolumn{4}{c}{mAP \%}\\\cline{3-6} 
& & $k=1$ & $k=3$ & $k=5$ & $k=All$\\\hline
Knowledge Gating & WBF &76.30 &75.56 &74.66 &73.96\\
Knowledge Gating & NMS &\textbf{77.59} &76.37 &76.53 &75.81\\
Knowledge Gating & Soft-NMS &76.95 &68.75 &68.75 &65.71\\\hline
Deep Gating & WBF &67.62 &75.19 &72.95 &73.96\\
Deep Gating & NMS &67.43 &\textbf{78.14} &73.31 &75.81\\
Deep Gating & Soft-NMS &67.27 &77.36 &74.70 &65.71\\\hline
Attn. Gating & WBF &67.86 & 74.54 &72.93 &73.96\\
Attn. Gating & NMS &67.43 & 78.51 &73.47 &75.81\\
Attn. Gating & Soft-NMS &67.27 & \textbf{81.31} &69.88 &65.71\\\hline
Optimal Gating & WBF &75.57 & 74.62 &72.45 &73.96\\
Optimal Gating & NMS &74.69 & 77.10 &73.20 &75.81\\
Optimal Gating & Soft-NMS &73.03 & \textbf{81.57} &72.93 &65.71\\\hline
\end{tabular}
\caption{Evaluation of different gating methods for selecting the $k$ best branches for across different fusion methods. For each input, the top $k$ branches selected by the gate are fused to produce a set of detections scored using mAP.}
\label{tab:gating_eval_extended}
\vspace{-5mm}
\end{table}

\begin{table*}[!h]
    \centering
    \begin{tabular}{c c c c c c c c c}
    \hline
        \multirow{2}{*}{$k$} & \multirow{2}{*}{Gate Model} & \multicolumn{7}{c}{Branch Selection Rate (\%)} \\\cline{3-9}
        & & Radar& L Cam.& R Cam.&Lidar&L/R Cam. &L/R Cam.+ Lidar& Radar+Lidar\\\hline
        \multirow{4}{*}{1} 
&KnowledgeGating    & 8.61 & 0.00 & 0.00 & 0.00 & \textbf{76.82} & 0.00 & 14.57\\
&DeepGating         & 7.95 & 0.00 & 0.00 & 18.54 & 1.32 & 19.87 & \textbf{52.32}\\
&AttentionGating    & 9.93 & 0.66 & 0.00 & 6.62 & 8.61 & 12.58 & \textbf{61.59}\\
&Optimal Gating     & 19.87 & 4.64 & 1.99 & \textbf{25.83} & 13.25 & 9.27 & 25.17\\\hline
        \multirow{4}{*}{3} 
&KnowledgeGating    & 23.18 & \textbf{76.82} & \textbf{76.82} & 23.18 & \textbf{76.82} & 0.00 & 23.18\\
&DeepGating         & 78.81 & 3.31 & 12.58 & 66.23 & 25.17 & 31.79 & \textbf{82.12}\\
&AttentionGating    & 78.81 & 1.32 & 7.95 & 72.19 & 28.48 & 28.48 & \textbf{82.78}\\
&Optimal Gating     & 74.17 & 16.56 & 15.23 & 62.91 & 32.45 & 23.84 & \textbf{74.83}\\\hline
        \multirow{4}{*}{5} 
&KnowledgeGating    & 23.18 & 76.82 & 76.82 & 23.18 & \textbf{100.00} & \textbf{100.00} & \textbf{100.00}\\
&DeepGating         & \textbf{97.35} & 32.45 & 21.19 & 87.42 & 82.78 & 94.04 & 84.77\\
&AttentionGating    & 89.40 & 29.14 & 19.87 & \textbf{96.03} & 76.82 & 95.36 & 93.38\\
&Optimal Gating     & \textbf{87.42} & 54.30 & 47.68 & 80.13 & 67.55 & 78.81 & 84.11\\\hline
    \end{tabular}
    \caption{Evaluation on how often each gate model selected each branch as part of the top-$k$ for various $k$. The branch selection rate is expressed as a percentage over the number of inputs for the test dataset.}
    \label{tab:branch}
\end{table*}

Table \ref{tab:gating_eval_extended} shows our extended gating results. It includes mAP scores evaluated with the four gating strategies, each used with WBF, NMS, and Soft-NMS fusion. The results indicate that NMS and Soft-NMS with $k=3$ result in the highest mAP score for most of the gating methods evaluated. Interestingly, NMS seems to be more robust to different $k$ values than Soft-NMS, which varies by up to 15\% depending on $k$. WBF works well with Knowledge Gating for all $k$ and works decently well for the other gates with $k \in \{3,5,All\}$. As mentioned in Section \ref{gating_evaluation}, regardless of the performance of other configurations, only the highest scoring configuration would be deployed in an actual vehicle. Thus, these extended results confirm that Attention Gating with Soft-NMS and $k=3$ is the best configuration to deploy in the real world.

\subsection{Branch Selection}

In Table \ref{tab:branch}, we show the frequency at which each branch was selected by each gate model for different values of $k$. The branches listed are the seven branches explicitly defined in our experiments (four single-sensor and three early fusion). The branch selection rate is the percent of inputs in the test dataset for which a specific branch was selected as part of the top-$k$. The selection results for a single input can vary depending on the context; however, these aggregated results illuminate which sensors contributed more to the final detection results than others. The deep learning-based gating models heavily favored selecting radar and lidar branches. We propose that this dependence on sensors that are traditionally more robust to severe weather was reinforced throughout the model's learning process as feedback taught the model that the cameras were susceptible to high amounts of error in specific contexts. Knowledge gating tends to favor the camera selection more often but does not perform as well as the deep learning models (see Table \ref{tab:gating_eval_extended}). This illuminates a limitation in using domain knowledge to define the gate as some sensors, like cameras, dominate the selection process. Optimal Gating shows the most consistent responses across the branches as expected due to its \textit{a posteriori} knowledge of each branch's loss.

\end{document}